\pdfoutput=1
\documentclass[11pt,twocolumn]{article}

\usepackage{geometry}
\usepackage{authblk}

\usepackage{times}
\usepackage{latexsym}

\usepackage[T1]{fontenc}
\usepackage[utf8]{inputenc}

\usepackage{microtype}

\usepackage{inconsolata}

\usepackage{booktabs}
\usepackage{subcaption}
\usepackage{amsfonts}
\usepackage{amsmath}
\usepackage{multirow}
\usepackage{spverbatim}
\usepackage{float}
\usepackage{xcolor}

\usepackage{graphicx}
\usepackage{enumitem}
\usepackage{subfig}

\usepackage{graphicx}

\usepackage{url}
\usepackage{verbatim}
\usepackage{natbib}
\title{Improving Score Reliability of Multiple Choice Benchmarks with Consistency Evaluation and Altered Answer Choices}

\author[1]{Paulo Cavalin}
\author[1]{Cassia Sanctos}
\author[1]{Marcelo Grave}
\author[1]{Claudio Pinhanez}
\author[1]{Yago Primerano}
\affil[1]{IBM Research Brazil}

\begin{document}
\maketitle
\begin{abstract}
    In this work we present the Consistency-Rebalanced Accuracy (CoRA) metric, improving the reliability of Large Language Model (LLM) scores computed on multiple choice (MC) benchmarks. Our metric explores the response consistency of the LLMs, taking advantage of synthetically-generated questions with altered answer choices. With two intermediate scores, i.e. Bare-Minimum-Consistency Accuracy (BMCA) and Consistency Index (CI), CoRA is computed by adjusting the multiple-choice question answering (MCQA) scores to better reflect the level of consistency of the LLM. We present evaluations in different benchmarks using diverse LLMs, and not only demonstrate that LLMs can present low response consistency even when they present high MCQA scores, but also that CoRA can successfully scale down the scores of inconsistent models.

\end{abstract}

\section{Introduction}
Despite the current popularity of Large Language Models (LLMs), and the undeniable capabilities that they have demonstrated to solve very complex real-world problems, it is also the real truth that there is yet a lot to be done in terms of understanding and measuring precisely their capabilities and risks to deploy reliable and liable applications. 

The most used approach to evaluate LLMs is to measure its performance on question-answering (QA) benchmark datasets (or simply benchmarks), i.e. datasets containing questions (aka prompts) with their respective expected outputs, where the outputs generated by the LLM are compared to the expected outputs from the benchmark, resulting in univariate scores that are used to rank and evaluate the LLMs. A common way to structure QA benchmarks is to rely on multiple choice (MC) questions, which is not only a widely adopted method to evaluate human for several knowledge-testing objectives, but also has the advantage of being a very simple way to compute right and wrong answers.

Although the research community has been highly active in investigating the limitations of current benchmarking practices, and several issues have already been identified for MC evaluations, such as choice biases, variability to rewordings, inconsistent confidence, among others \citep{zheng2024largelanguagemodelsrobust,reif-schwartz-2024-beyond,ye2024benchmarkingllmsuncertaintyquantification}, we believe that there is still a gap in better quantifying the capabilities of an LLM. Given that the most used method to evaluate LLM on MC benchmarks is to compute the ratio of matches between the responses of the LLM against the correct alternatives, an approach that we refer to MCQA, we think that this approach is lacking in providing a realistic and informative evaluation whether the LLM is actually knowledgeable about the test questions, or if that the scores are a by-product of issues such as training data contamination or random guesses given the stochastic nature of inference algorithms.

In this work we argue that computing response consistency is key to have metrics that are able to present a more reliable score for the evaluation of LLMs on MC benchmarks. As already demonstrated, LLMs can suffer from  inconsistencies when subjected to variations in the input, especially when the set of presented alternatives is slightly modified with reorderings or changes in the set of distractors \citep{pezeshkpour-hruschka-2024-large,wang-etal-2025-llms-may}. Notice that distractors consists of the alternatives in a MC question that are not correct, so usually a MC question is composed with a question and a set of choices, where there is one correct choice\footnote{Although it is possible to have more than correct choice in a MC question, we delimit the scope of this work for cases with only one correct alternative.} and one or more distractors. Thus, it is quite easy to synthetically generated altered questions by playing with the set of distractors, while keeping the correct choice.

Based on generating altered sets of questions with modified distractors (or simple reorderings) in the choices, a set to which we refer as the divergent questions, we propose the Consistency-Rebalanced Accuracy (CoRA) metric to better reflect the level of consistency of the LLM on the MC benchmark. The metric is based on two intermediate metrics, i.e. Bare-Minimum-Consistency Accuracy (BMCA) and Consistency Index (CI), where the first is used to compute the accuracy according to a specified minimum level of consistency, and the second computes the gap of accuracy between the score on the original benchmark, i.e. the MCQA method, and the BMCA(1.0), the accuracy for 100\% consistency. The CoRA score is then computed by scaling the related MCQA score with the value computed with CI.

We evaluate CoRA in different popular benchmarks, with both open source and a commercial LLM, and observe that CoRA tends to reflect more realistic  distribution of scores according to the consistency level of the LLMs. That is, with BMCA evaluated with different levels of minimum consistency, we observe that some top-performing LLMs present a drastic decrease of accuracy, and reach very poor performances with BMCA(1.0), indicating that the scores with MCQA and the other baselines are not reliably reflecting the consistency of the LLM. Consequently, scaling down that score with the consistency index CI results in a more faithful measurement of the capabilities of the LLM: the CoRA metric.

We believe that this paper not only contributes to improve the evaluation of LLMs in MC benchmarks, but also in emphasizing that the use of response consistency evaluation is a viable and necessary approach to provide more faithful benchmark evaluation scores. We show an LLM such as GPT4o can present a drop of at least 0.10 points in accuracy, comparing MCQA with BMCA(1.0), showing that even this top-performing LLM can be `unsure' about the correct answer for about 14\% of the correct response. More concerning is that models that perform close to GPT4o in MCQA score, such as MedLlama3, can present very low consistency levels, making the CoRA score to be less than half of the original MCQA score. In our opinion, it is mandatory to include consistency evaluation before releasing any score computed on an MC benchmark.

In order to make our research accessible by the community, we are publicly releasing the source code for computing CoRA scores: \url{https://github.com/IBM/cat}.

% Although previous metrics have already tackled somehow the use of consistency response, such as the MCQA+ and MV metrics, we show that they fall short in penalizing LLMs with low consistency, usually keeping the derived scores close to the values computed with MCQA.

\section{Related Work}
Understanding well the capabilities of LLMs is key for deploying safe, liable, end-user applications, and several efforts have been made towards improving the evaluation of such models \citep{lin-chen-2023-llm,wang-etal-2024-assessing,lei-etal-2024-s3eval}. Although we can see some works focusing on the evaluation of open-end questions \citep{myrzakhan2024openllmleaderboardmultichoiceopenstylequestions}, multiple-choice (MC) evaluation is a common practice for mainstream models \citep{singhal2023expertlevelmedicalquestionanswering,jiang2023mistral7b,nori2023generalistfoundationmodelsoutcompete,dubey2024llama3herdmodels}. Multiple-choice questions can be more objectively evaluated as opposed to open questions, the evaluation for which can be difficult and subjective even for human evaluators. Furthermore, MC evaluation is a widely-used practice to evaluate proficiency of humans in several areas, for instance medical and law domains \citep{lesage2013188,curtis2013,grazziotin-soares2021}. It is thus natural to rely on a similar evaluation process to measure the proficiency of LLMs.

It is well known, though, that there is room to make MC evaluation more reliable and believable. Some efforts have been made in trying to understand the limitations of MC evaluation focused on confidence levels, either considering the logits of the neural networks or self-confidence scores provides by the LLM itself \citep{ye2024benchmarkingllmsuncertaintyquantification}. In \citep{wiegreffe2024answerassembleaceunderstanding}, an analysis on how the weights of transformers react to predict a correct answer is presented. The correlation between model confidence (probability outputs) and model self-confidence (a confidence level expressed by the model) have explored in \citep{kumar-etal-2024-confidence}, which show that the LLMs usually present low to moderate correlation. Others have focused on identifying possible biases that can be exhibited by the LLM, such as selection, token, and label biases \citep{zheng2024largelanguagemodelsrobust,reif-schwartz-2024-beyond}. 

Another group of researchers focused on understanding the sensibility of the LLM according to changes in the input. In \citep{mirzadeh2024gsmsymbolicunderstandinglimitationsmathematical}, the authors show that LLMs are negatively impacted by changes in the input question. In this case, the model performs significantly worse when only the numbers are changed in the input for math-related questions. Assuming that LLMs can be affect by changes in the input, the work presented in \citep{ackerman-etal-2024-novel} proposes a metric for computing the robustness of LLMs to input changes, considering perturbations in the input and reporting the impact in the accuracy. Intriguing results were reported in \citep{balepur-etal-2024-artifacts}, where the authors query LLMs only with choices without the question was investigated, and show that even without the questions the LLMs can correctly answer a considerable number of questions. The authors looked for memorization but could not fully explain the phenomenon.

One particular line of research focuses in investigating the \emph{consistency} of LLMs in providing a response when the question is kept intact but with variations in other factors, such as the set of choices and parameters of the inference algorithm. An investigation on the sensitivity of choice order is presented \citep{li-etal-2024-multiple}, along with an exploration on the consistency of LLM according to different values for the temperature parameter, but they show that models such as GPT-3.5 tend to present high-consistency when prompted with different temperature values. In \citep{wei2024rethinkinggenerativelargelanguage}, the authors compare the results of MC evaluation and open-ended answer, and find low consistency between these two evaluations. However, both works presented in \citep{pezeshkpour-hruschka-2024-large} and \citep{wang2024answersreviewingrationalitymultiple} explore changes in the set of choices, either with simple reorderings or by modifying the set of distractors, and provide convincing evidence that LLMs that perform well is some MC benchmark are prone to lack of consistency when facing questions with modified distractors. As a consequence, two new metrics exploiting the consistency of LLM, namely MCQA+ and MV (see Section~\ref{sec:baselines}), have been proposed.

This work is heavily inspired by the results on the consistency of LLM when the sets of choices are modified. Our main contribution is on improving the robustness of metrics to more faithfully take into account the consistency level of a given LLM on a MC benchmark and express that into a score.

\section{Baselines}
\label{sec:baselines}
In this section we will revisit how the accuracy score is computed for MC benchmarks. We will first describe how this is done in the most traditional method, i.e. to perform single-run evaluations and compute the ratio of hits. Next, we will also describe existing methods that explore divergent sets of answers to enhance the computation of such scores. 

Let $MCQ = \{mcq_1, \ldots, mcq_N\}$ be the original set of $N$ questions, choice, and answers of a MC benchmark. Consider also that there is a function denoted $llm(mcq_i)$ that returns 1 if a given LLM presents the correct response, i.e. the response provided by the LLM is equal to the correct alternative in $mcq_i$, or 0 otherwise. The most used baseline consists of computing the accuracy directly on $MCQ$, which we refer to as MCQA and define as: 
\begin{equation}
    \mbox{MCQA} = \frac{1}{N} \displaystyle\sum_{i=1}^N llm(mcq_i)
\end{equation}

But as we presented in the previous section, LLMs can be inconsistent even we very simple test, and it is important to take such aspect into account during the evaluation process. As a consequence, consider also that there is a set denoted $MCQ* = \{ \hat{MCQ*}_1, \ldots, \hat{MCQ*}_N \}$, comprising $N$ divergence sets which are derivations of the samples in the $MCQ$ set. The divergence sets can be created using $M$ different methods, so that $\hat{MCQ*}_i = \{mcq*^1_1, \ldots, mcq*^M_i\}$ and $mcq_i \in \hat{MCQ*}_i$, i.e. the original question can also be included in the divergence set. We can find in the literature some methods that explore the of creating divergence sets and using them to materialize into metrics \citep{pezeshkpour-hruschka-2024-large,wang-etal-2025-llms-may}. 

One metric is MCQA+, based on the idea of computing the MCQA scores using disjoint divergence sets and aggregating the results using the mean of all $M$ divergence sets, i.e. the mean for the entire set of questions. This metric can be defined as:
\begin{equation}
    \mbox{MCQA}+ = \frac{1}{N*M} \displaystyle\sum_{i=1}^N \displaystyle\sum_{j=1}^M llm({mcq*}^j_i)
\end{equation}

Notice that MCQA+ considers the divergence sets for generating alternative evaluations, but the aggregation of the scores under-explores the computation of consistency. In fact, we can say that MCQA+ does rely on an implicit use consistency, but given that incorrect response also contribute to the score, it is not trivial to associate the metic to the consistency of correct responses only.

Another metric that includes consistency in the computation of accuracy scores is the Majority Voting (MV) metric, proposed in \citep{pezeshkpour-hruschka-2024-large}. This metric relies on the set of the divergent sets $MCQ*$ and computes the correctness of an evaluation sample based on achieving the correct response in the majority of the derivations, i.e. if the LLM provides the correct response for more than half of the samples in $\hat{MCQ*}_i$, or in other words, more than half response consistency. That is, consider the definition response consistency for a given sample $i$ as:
\begin{equation}
    \mbox{RC(i)} = \frac{1}{M} \displaystyle\sum_{j=1}^{M} llm(mcq*^j_i)
\end{equation}

Now, consider the function $1(expression)$, which returns 1 if $expression$ is true or 0 otherwise, the MV metric can then be defined as:
\begin{equation}
    \mbox{MV} = \frac{1}{N} \displaystyle\sum_{i=1}^{N} 1(\mbox{RC(i)} > 0.5)
\end{equation}

% To follow a more uniform notation, we will refer to MV as MCQA+MV hereafter, using the plus sign to represent an extended version of MCQA with divergent questions, and that MV is the aggregation method.

Even though MV relies on consistency to compute hits, i.e. the majority of the divergent questions need to get correct responses for a question to be computed as a hit, the metric relies on very permissive level of consistency (0.5), for which samples with low response consistency values have the same weights of those with higher values. In some sense, that hinders the impact of consistency in the metric, apart from being a more statistically-robust score compared with MCQA.

\section{Using Consistency for More Reliable Accuracy Computation}
In this section we propose our method to rebalance LLM accuracy on MC benchmarks, to which we refer as the Consistency-Rebalanced Accuracy (CoRA). The metric is built upon the idea of computing the Consistency Index (CI) score using the score computed with the Bare-Minimum-Consistency Accuracy (BMCA) method using 100\% consistency as target, and then adjusting the scores computed with the MCQA method to scale down LLMs that are inconsistent, resulting in CoRA scores. Details are provided next.

\subsection{Bare-Minimum-Consistency Accuracy}
The first metric we propose is the Bare-Minimum Consistency Accuracy (BMCA), which can be considered as an extension of the MV metric but using the adjustable parameter $c$ to determine the minimum response consistency level that is expected for a sample to be considered correct. That is, for each sample $i$, we will only consider the samples as correct if the RC score is greater than $c$

In greater details, given the consistency level $c$ as a parameter, the BMCA metric can be defined as:
\begin{equation}
    \mbox{BMCA(c)} = \frac{1}{N} \displaystyle\sum_{i=1}^{N} 1(\mbox{RC}(i) \ge c)
\end{equation}

\subsection{Consistency Index}
Given that BMCA can compute scores for different levels of consistency, when $c = 1.0$, the metric will compute the accuracy score only for cases where the model provide 100\% response consistency in the M trials.

We associate here the idea of this index with the elimination of random guessing as a viable option for the models being evaluated. As detailed in the appendix~\ref{app:guessing_benchmarks}, for $M = 10$, a model has to be guessing at a success rate greater than 0.9999 to be able to be 100\% consistent on $M$ trials. When no random guess is allowed at all, the LLM is arguably knowledgeable about the responses provided for the benchmark questions. 

Therefore, we use BMCA(1.0) as a proxy to define the proportion of samples for which the LLM being evaluated demonstrates real knowledge when answering the questions, and use the score to compute a quality metric for the original MCQA score. We refer to this metric as the Consistency Index (CI) score.

Formally, the CI score is computed using the difference between MCQA and BMCA(1.0), providing the gap of the MCQA score to the accuracy considering only cases with 100\% of response consistency, and subtracting from $1.0$ so a higher value denotes higher consistency, such as in:
\begin{equation}
    \mbox{CI} = 1.0 - (\mbox{MCQA} - \mbox{BMCA(1.0)})
\end{equation}

\subsection{Consistency-Rebalanced Accuracy}
The end result of our approach is the Consistency-Rebalanced Accuracy (CoRA), consisting of scaling down the scores computed with MCQA using the CI score described in the previous section. The idea is to take advantage of the CI score and adjust MCQA scores to make them reflect more authentically the quality of the LLM in terms of response consistency.

The implementation is straightforward, as we denote in the equation below:
\begin{equation}
    \mbox{CoRA} = \mbox{MCQA} * \mbox{CI}
\end{equation}

With this approach, the MCQA scores are at best kept, if the model presents 100\% response consistency for the correct responses (which is very unlikely, as we will show later), but they can be scaled down as the LLM scores presents larger gaps from BMCA(1.0) to MCQA.

\begin{figure*}[t]
    \centering
    \includegraphics[width=.85\textwidth]{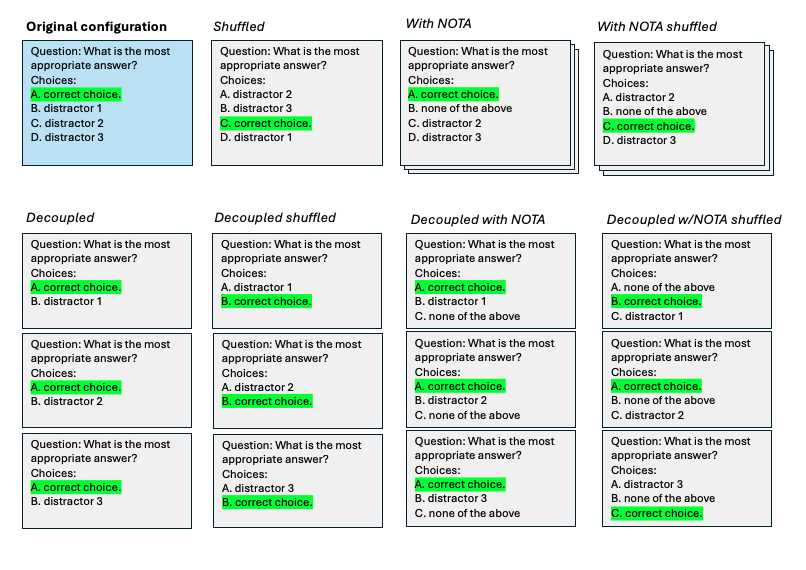}
    \caption{Illustration of the methods to create divergent sets of alternatives}
    \label{fig:jumbling_methods}
\end{figure*}

\section{Details for Implementation}
\label{sec:variations}
As a method to implement the divergent set $MCQ*$, we focus in creating derivations of the set of choices $C$ by creating variations only in the way multiple options are presented to the model. Although we can also vary the set of questions $Q$ with the variations in the input, such as with rephrasings, we wanted to avoid introducing any potential error in this process. By including variations in the set $C$, we can generate alternative sets of choices where the expected answer is always kept but the set of distractors is modified with three operators: reorderings, where the order of the choices presented to the LLM is modified, such as by shuffling the choices; deletions, where one or more alternatives are removed from the original set of choices; and inclusions, where alternatives that do not affect the expected answers are included, such as adding a \emph{none of the above} (NOTA) alternative.

In details, consider that $A$ is the number of alternatives in the original question, we consider the following approaches to generate altered sets of answer choices: {\bf Shuffled}, where the set of choices is shuffled (we can shuffle multiple times, but in this work we shuffle only once); {\bf With NOTA}, where each distractor is replaced by the NOTA alternative, resulting in $A-1$ new sets of alternatives; {\bf With NOTA shuffled}, employing a mix of \emph{With NOTA} and \emph{Shuffled}, where the NOTA alternative replaces a distractor and them the set of choices is shuffled (this approach also results in $A-1$ variations); {\bf Decoupled}, which takes the original set of choices and decouples it into $A-1$ binary subset of choices, pairing each of non-correct choices with the correct one in each subset; {\bf Decoupled shuffled}, similar to \emph{Decoupled} but with an additional shuffling step; {\bf Decoupled with NOTA}, also mixing \emph{Decoupled} with \emph{With NOTA}, where the set of alternatives is decoupled into $A-1$ binary subsets and a NOTA distractor is add to each subset, creating ternary subsets; {\bf Decoupled with NOTA shuffled}, which is similar to \emph{Decoupled with NOTA} with additional shufflings on the ternary subsets.

An illustration of the previously described techniques is presented in Figure~\ref{fig:jumbling_methods}. With these variations in the set of choices, we can then format a prompt for each variation and prompted the model for the correct alternative. We consider the following base prompt, where \$QUESTION\$ is replaced by the text of the question, followed by the corresponding options which are formatted and filled in \$CHOICES\$:
\begin{small}
\begin{verbatim}
    Answer the following multiple choice question. 
    The first line of your response should be of 
    the following format: 'LETTER' (without 
    quotes), where LETTER is one of ABCD 
    (depending on the number of alternatives), 
    followed by a step-by-step explanation.

    Question: $QUESTION$
    Choices: $CHOICES$
    Answer:
\end{verbatim}
\end{small}

To evaluate the output of the LLMs, we parse the first token of the response and remove any additional character such as punctuation and spaces. 

Our method generates a total of $2+6*(A-1)$ variations of the set of choices for each question. For instance, with with five alternatives, i.e. $A=5$, 26 variations are generated. Notice that one could easily generate more variations by including more exhaustive shufflings and generate subsets with more than two and less than M alternatives, but that implies in extra costs to evaluate the models since we need to query the model for each variations of the original question, and there is a linear increase in the cost of running LLM inferences.

\section{Empirical Evaluation}
In this section we describe the evaluations conducted to validate our proposed CoRA metric. We focus on comparing the results against the three baselines described in Section~\ref{sec:baselines}: MCQA, MCQA+, and MCQA+MV.

We divide this evaluation into two main parts. In this first we focus on a domain-specific dataset, i.e. the MedQA benchmark \citep{jin2020disease}, considering four different LLMs either finetuned on medical data or known as good performer in this type of data: GPT4o version "gpt-4o-2024-11-20", MedLlama3 7B (MedL) \citep{ProbeMedicalYonseiMAILab-medllama3-v20}, BioMedical Llama3 8B (BioML) \citep{ProbeMedicalYonseiMAILab-medllama3-v20}, and BioMistral 7B (BMist) \citep{labrak2024biomistral}. In the second part we evaluate three general knowledge benchmarks, i.e. MMLU \citep{hendryckstest2021}, Arc-C \citep{allenai:arc}, and TruthFulQA \citep{lin-etal-2022-truthfulqa}, considering four general-purpose LLMs: Mistral v0.1 7B (Mist) \citep{jiang2023mistral7b}, Llama 3.1 8B (Llam) \citep{grattafiori2024llama3herdmodels}, Granite 3.0 8B (Gran) \citep{granite2024granite}, and DeepSeek chat 7B (DSeek) \citep{deepseek-chat}. Notice that we focused on models with slightly similar sizes, i.e. around 7B to 8B parameters, except for GPT4o whose size is not publicly disclosed. For MedQA we consider the 1,273 5-choice questions extracted from USMLE exams, resulting in generating 26 variations for the $MCQ*$ set of divergent questions, and use 0-shot prompts. For MMLU, we used 14,042 questions with either 3 or 4 alternatives, generating 14 to 20 questions in $MCQ*$, and evaluated with 5-shot prompts. For Arc-C, we used 1,172 questions with either 4 or 5 alternatives, generating 20 to 26 questions in $MCQ*$, and evaluated with 25-shot prompts. And for TruthFulQA, we consider 817 questions ranging from 2 or 12 alternatives, generating 14 to 74 questions in $MCQ*$, and evaluated with 0-shot prompts. Notice that the decision for in-context learning method was based on common practices from the literature \citep{grattafiori2024llama3herdmodels,deepseekai2024deepseekv3technicalreport}.

For a better understanding of the CoRA scores, for each LLM and dataset we present an extensive evaluation of BMCA considering six different values for $c$, i.e. 0.5, 0.6, 0.7, 0.8, 0.9, and 1.0, noticing that 0.5 represents a borderline level consistency for deciding if a question is correctly responded or not, and 1.0 corresponds to full consistency, i.e. all responses from the divergent set of questions are correct. That range of values is useful to provide a progressive analysis on the consistency of an LLM, i.e. how the accuracy is affect as we increase the value of $c$. We report also all values computed for the CI score.

\begin{table}[htb]
    \centering
    \caption{Results on MedQA dataset. In parentheses the comparison ranking.} 
    \label{tab:results_medqa}
    \begin{footnotesize}
    \begin{tabular}{l|cccc}
    \hline
    LLMs:  & GPT4o & MedL & BioML & BMist \\
    \hline
    \multicolumn{5}{c}{Baselines} \\ 
    \hline
    \textbf{MCQA}   & 0.85 (1) & 0.74 (2) & 0.73 (3) & 0.38 (4) \\
    \textbf{MCQA+}  & 0.90 (1) & 0.69 (3) & 0.75 (2) & 0.58 (4) \\
%    \textbf{MCQA+($\times 1$})  & - & - & - & - \\
    \textbf{MV}  & 0.91 (1) & 0.73 (3) & 0.77 (2) & 0.58 (4) \\    
    \hline
    \multicolumn{5}{c}{Proposed metric} \\
    \hline
    \textbf{CoRA}     & 0.74 (1) & 0.32 (3) & 0.42 (2) & 0.28 (4) \\    
    \hline
    \multicolumn{5}{c}{Secondary metrics} \\
    \hline
    \multicolumn{2}{l}{\textbf{BMCA(c)}} & & & \\
    \textbf{$c \ge 0.5$} & 0.91 & 0.75 & 0.80 & 0.61 \\
    \textbf{$c \ge 0.6$} & 0.89 & 0.67 & 0.73 & 0.49 \\
    \textbf{$c \ge 0.7$} & 0.86 & 0.57 & 0.66 & 0.40 \\
    \textbf{$c \ge 0.8$} & 0.84 & 0.49 & 0.60 & 0.33 \\
    \textbf{$c \ge 0.9$} & 0.79 & 0.34 & 0.46 & 0.22 \\
    \textbf{$c \ge 1.0$} & 0.73 & 0.18 & 0.31 & 0.11 \\
    \hline
    \textbf{CI}     & 0.88 & 0.44 & 0.58 & 0.73 \\
    \hline
    \end{tabular}
    \end{footnotesize}
\end{table}

The results on MedQA are presented in Table~\ref{tab:results_medqa}. When comparing the scores of MCQA+ and MV baselines to MCQA, it is noticeable how the scores of all models, except MedL, are magnified with the expanded evaluation with divergent questions. On the other hand CoRA, presents reduced scores by about 0.11 points for GPT4, 0.42 points for MedL, 0.31 for BioML, 0.10 for BMist. Note that such a decrease is proportional to the CI score, reported in the last row of the table. By looking at the scores from BMCA(1.0) and comparing them to those of MCQA, we can clearly observe that there is a drop for all models, being GPT4o undeniably more consistency, reaching a score of 0.73, i.e. 73\% of the correct responses present $RC(i) = 1.0$. For the other models, on the other hand, less than half of the correct cases present 100\% response consistency, and that gap in consistency is represented in CoRA scores that are much lower compared with MCQA. That is, MedL drops from 0.74 with MCQA to 0.32 with CorA, BioML from 0.73 to 0.42, and BMist from 0.38 to 0.28. Further evidence on the differences in consistency can be found by looking at the different scores provided by BMCA, showing that models such as MedL and BioML present a drastic decrease in accuracy as the minimum consistency requirement increase, while the decrease is not as drastic for BioML and very subtle for GPT4.

\begin{table}[htb]   
    \centering
    \caption{Results on MMLU dataset. In parentheses the comparison ranking.} 
    \label{tab:results_mmlu}
    \begin{footnotesize}
    \begin{tabular}{l|cccc}
    \hline
    LLMs:  & Mist & Llam & Gran & DSeek \\
    \hline
    \multicolumn{5}{c}{Baselines} \\ 
    \hline
    \textbf{MCQA}   & 0.64 (2) & 0.58 (3) & 0.65 (1) & 0.52 (4) \\
    \textbf{MCQA+}  & 0.75 (1) & 0.62 (3) & 0.75 (1) & 0.60 (4) \\
    % \textbf{MCQA+($\times 1$})  & -    & -    & -    & -    \\
    \textbf{MV} & 0.78 (1)   & 0.61 (3)  & 0.76 (2)  & 0.58 (4)   \\    
    \hline
    \multicolumn{5}{c}{Proposed metric} \\
    \hline
    \textbf{CoRA}     & 0.44 (2) & 0.33 (3) & 0.47 (1) & 0.33 (3) \\    
    \hline
    \multicolumn{5}{c}{Secondary metrics} \\
    \hline
    \multicolumn{2}{l}{\textbf{BMCA(c)}} & & & \\
    $c \ge 0.5$     & 0.83 & 0.65 & 0.80 & 0.63 \\
    $c \ge 0.6$     & 0.74 & 0.55 & 0.73 & 0.54 \\
    $c \ge 0.7$     & 0.65 & 0.46 & 0.66 & 0.44 \\
    $c \ge 0.8$     & 0.56 & 0.38 & 0.58 & 0.36 \\
    $c \ge 0.9$     & 0.47 & 0.28 & 0.50 & 0.27 \\
    $c \ge 1.0$     & 0.33 & 0.15 & 0.38 & 0.16 \\
    \hline
%    \textbf{EC}     & 0.7  & 0.6  & 0.7 & 0.6  \\
    \textbf{CI}     & 0.69 & 0.57 & 0.73 & 0.64 \\
    \hline
    \end{tabular}
    \end{footnotesize}
\end{table}

\begin{table}[htb]   
    \centering
    \caption{Results on Arc-C dataset. In parentheses the comparison ranking.} 
    \label{tab:results_arc}
    \begin{footnotesize}
    \begin{tabular}{l|cccc}
    \hline
    LLMs:  & Mist & Llam & Gran & DSeek \\
    \hline
    \multicolumn{5}{c}{Baselines} \\ 
    \hline
    \textbf{MCQA}   & 0.80 (2) & 0.72 (3) & 0.82 (1) & 0.64 (4) \\
    \textbf{MCQA+}  & 0.90 (1) & 0.73 (3) & 0.89 (2) & 0.73 (3) \\
%    \textbf{MCQA+($\times 1$})  & -    & -    & -    & -    \\
    \textbf{MV} & 0.92 (1) & 0.81 (3) & 0.91 (2) & 0.78 (4) \\    
    \hline
    \multicolumn{5}{c}{Proposed metric} \\
    \hline
    \textbf{CoRA} & 0.58 (2) & 0.25 (4) & 0.65 (1) & 0.38 (3) \\    
    \hline
    \multicolumn{5}{c}{Secondary metrics} \\
    \hline
    \multicolumn{2}{l}{\textbf{BMCA(c)}} & & & \\
    $c \ge 0.5$     & 0.94 & 0.84 & 0.92 & 0.82 \\
    $c \ge 0.6$     & 0.90 & 0.78 & 0.88 & 0.73 \\
    $c \ge 0.7$     & 0.84 & 0.69 & 0.84 & 0.64 \\
    $c \ge 0.8$     & 0.80 & 0.53 & 0.80 & 0.53 \\
    $c \ge 0.9$     & 0.70 & 0.32 & 0.73 & 0.42 \\
    $c \ge 1.0$     & 0.52 & 0.07 & 0.61 & 0.23 \\
    \hline
%    \textbf{EC}     & 0.8  & 0.7  & 0.8  & 0.7 \\
    \textbf{CI}     & 0.72 & 0.35 & 0.79 & 0.59 \\
    \hline
    \end{tabular}
    \end{footnotesize}
\end{table}

\begin{table}[htb]
    \centering
    \caption{Results on TruthfulQA dataset. In parentheses the comparison ranking.} 
    \label{tab:results_truth}
    \begin{footnotesize}
    \begin{tabular}{l|cccc}
    \hline
    LLMs:  & Mist & Llam & Gran & DSeek \\
    \hline
    \multicolumn{5}{c}{Baselines} \\ 
    \hline
    \textbf{MCQA}   & 0.41 (1) & 0.41 (1) & 0.34 (3) & 0.34 (3) \\
    \textbf{MCQA+}  & 0.49 (1) & 0.49 (1) & 0.39 (4) & 0.40 (3) \\
    %\textbf{MCQA+($\times 1$})  & - & - & - & - \\
    \textbf{MV}  & 0.45 (2) & 0.47 (1) & 0.35 (4) & 0.37 (3) \\
    \hline
    \multicolumn{5}{c}{Proposed metric} \\
    \hline
    \textbf{CoRA}     & 0.28 (1) & 0.27 (2) & 0.25 (3) & 0.24 (4) \\    
    \hline
    \multicolumn{5}{c}{Secondary metrics} \\
    \hline
    \multicolumn{2}{l}{\textbf{BMCA(c)}} & & & \\
    $c \ge 0.5$     & 0.48 & 0.51 & 0.36 & 0.40 \\
    $c \ge 0.6$     & 0.40 & 0.40 & 0.30 & 0.31 \\
    $c \ge 0.7$     & 0.33 & 0.31 & 0.25 & 0.23 \\
    $c \ge 0.8$     & 0.25 & 0.22 & 0.19 & 0.17 \\
    $c \ge 0.9$     & 0.16 & 0.14 & 0.14 & 0.11 \\
    $c \ge 1.0$     & 0.09 & 0.08 & 0.09 & 0.05 \\
    \hline
%    \textbf{EC}     & 0.6  & 0.6  & 0.5  & 0.6 \\
    \textbf{CI}     & 0.68 & 0.67 & 0.75 & 0.71 \\
    \hline
    \end{tabular}
    \end{footnotesize}
\end{table}

The results on general-knowledge benchmarks are presented in Table~\ref{tab:results_mmlu}, Table~\ref{tab:results_arc}, and Table~\ref{tab:results_truth}, for MMLU, Arc-C, and TruthfulQA, respectively. In this case, we observe a similar scenario apart the fact that there is not a very top-performing LLM such as GPT4o in the comparison. As it can be observed, the maximum CI score is of 0.79 in Arc-C with Gran, but CI can go as low as 0.35 with Llam in the same benchmark. That low CI score for Llam reflects a surprisingly inconsistency LLM on the Arc-C benchmark, and the resulting CoRA score reflects this lack of consistency. That is, from a difference of 0.10 MCQA points from Llam to Gran, the top performer in that benchmark, the difference becomes 0.40 points with CoRA. Notice that MCQA+ and MV keep gaps that are much closer to that of MCQA, i.e. 0.16 and 0.10, respectively. Overall, considering the best MCQA scores, that were achieved with Gran, we observe that our CoRA metric reflects at least a drop of 0.18 points in MMLU, 0.17 in Arc-C, and 0.09 in TruthfulQA. On the other hand, the drop in score for Llam, one of the worst performers in CoRA, can be as high as 0.25 points in MMLU, 0.57 in Arc-C, and 0.14 in TruthfulQA. It is intriguing that both the most and the least consistent LLM are the same for all three benchmarks, a fact that might indicate that the consistency might be a feature that generalizes among different benchmarks, but further investigation is needed to support this claim.

\subsection{Ablation Studies}
In this section we present an ablation study on the set of divergent questions used to compute our metrics. As depicted in Figure~\ref{fig:jumbling_methods}, some methods can keep the same number of alternatives as the original question (first row in Figure~\ref{fig:jumbling_methods}), but other methods can either augment or reduce that sets creating uneven uniform distribution with the likelihood of predicting the correct question, which can contribute for metrics such as MCQA+ and MV to produce higher scores. For this reason, in this section not only we present results with altered questions with only the exact same number of alternatives of the original question, but we also conduct a thorough statistical analysis on the set of altered questions based on bootstrap resampling. We focus on the MedQA dataset, which presents an invariable number of alternatives for the entire dataset and can be used for both evaluations.

Results with the set of only ten divergent questions with the same number of alternatives are presented in Table~\ref{tab:results_medqa_5-alternatives_only}. As somewhat expected, the results of MCQA+ and MV present a drop compared to the numbers reported in Table~\ref{tab:results_medqa}, the table containing analogous results with all 26 divergent questions. The score from CoRA, on the other hand, present slightly increases. We think that having a smaller set of divergent questions possibly reduces the impact of the consistency evaluation for these metrics, but we need further investigation to find more evidence to confirm this hypothesis.

\begin{table}[htb]
    \centering
    \caption{Results on MedQA dataset - 5-alternative questions only. In parentheses the comparison ranking.} 
    \label{tab:results_medqa_5-alternatives_only}
    \begin{footnotesize}
    \begin{tabular}{l|cccc}
    \hline
    LLMs:  & GPT4o & MedL & BioML & BMist \\
    \hline
    \multicolumn{5}{c}{5-alternative-only divergent questions} \\ 
    \hline
    \textbf{MCQA+}  & 0.86 (1) & 0.62 (3) & 0.67 (2) & 0.43 (4) \\
%    \textbf{MCQA+($\times 1$})  & - & - & - & - \\
    \textbf{MV} & 0.85 (1) & 0.62 (3) & 0.67 (2) & 0.38 (4) \\
    \textbf{CoRA}  & 0.77 (1) & 0.37 (3) & 0.48 (2) & 0.28 (4) \\    
    \hline
    \multicolumn{5}{c}{difference from Table~\ref{tab:results_medqa}} \\ 
    \hline
    \textbf{MCQA+}  & \textcolor{red}{-0.04} & \textcolor{red}{-0.07} & \textcolor{red}{-0.08} & \textcolor{red}{-0.15} \\
%    \textbf{MCQA+($\times 1$})  & - & - & - & - \\
    \textbf{MV}     & \textcolor{red}{-0.06} & \textcolor{red}{-0.11} & \textcolor{red}{-0.10} & \textcolor{red}{-0.20} \\    
    \textbf{CoRA}  & \textcolor{blue}{ 0.03} & \textcolor{blue}{ 0.05} & \textcolor{blue}{ 0.06} & 0.00 \\    
    \hline
    \end{tabular}
    \end{footnotesize}
\end{table}

We focus in understanding better the sensitiveness of the metric to the set of altered question, independently of the distribution and number of alternatives. For that we conducted a bootstrap resampling analysis, by generating 10,000 evaluations with 100 altered questions randomly selected with replacement from the 26 divergent questions created with MedQA.

The means of the scores computed with the 10,000 bootstrapped resamplings are presented in Table~\ref{tab:results_medqa_bootstrapped}. First, it is eye catching the usually low standard deviations, demonstrating an interesting stability of the score independently of the set of altered choices. Notice also that all methods present very small differences when compared to the seed divergence set with 26 variations, as reported in the bottom portion of the table. That is quite interesting since it indicates that the variations presented in Section~\ref{sec:variations} are relatively robust for the evaluation of response consistency, and our proposed method to generate variations can be used to evaluate LLMs without much concern with intrinsic variability from the set of altered choices.

\begin{table}[htb]
    \centering
    \caption{Results on MedQA dataset - means of 10,000 resamplings of 100-samples bootstrapped divergent questions. In parentheses the standard deviation multiplied by $10^3$.} 
    \label{tab:results_medqa_bootstrapped}
    \begin{footnotesize}
    \begin{tabular}{l|cccc}
    \hline
    LLMs:  & GPT4o & MedL & BioML & BMist \\
    \hline
    \multicolumn{5}{c}{10 bootstrapped divergent questions} \\ 
    \hline
    \textbf{MCQA+} & 0.90 (3) & 0.69 (11) & 0.75 (9)  & 0.58 (13) \\
    \textbf{MV}    & 0.91 (4) & 0.73 (15) & 0.78 (12) & 0.59 (21) \\
    \textbf{CoRA}  & 0.75 (1) & 0.33 ( 3) & 0.42 ( 2) & 0.28 ( 1) \\    
    \hline
    \multicolumn{5}{c}{difference to Table~\ref{tab:results_medqa}} \\ 
    \hline
    \textbf{MCQA+}  & 0.00 & 0.00 & 0.00 & 0.00 \\
    \textbf{MV}     & 0.00 & 0.00 & \textcolor{blue}{0.01} & \textcolor{blue}{0.01} \\    
    \textbf{CoRA}  & \textcolor{blue}{0.01} & \textcolor{blue}{0.01} & 0.00 & 0.00 \\    
    \hline
    \end{tabular}
    \end{footnotesize}
\end{table}

\section{Conclusions and Future Work}
In this work we proposed the CoRA metric to enhance the way LLMs are evaluated on MC benchmarks, which explores the concept of response consistency to rebalance the scores computed from the ratio for hits of an LLM on MC benchmarks and provide more faithfully the capabilities of such models. And our evaluations on well-known benchmarks show that CoRA is able to redistribute the scores according to the consistency of the LLM, which is demonstrated with the CI scores, improving the reliability of LLM evaluation compared to state-of-the-art metrics that do not reflect any aspect of response consistency in the scores. Furthermore, we conducted an ablation study focused at evaluating the sensitiveness of to the set of altered answer choices and demonstrate that our proposed generation method is relatively robust, practically equivalent to the scores obtained with bootstrap resampling.

As future work we believe we can improve the methodology in different ways. There is room to investigate and increase the set of divergent questions, and also in exploring further simpler and more general methods such as shuffling. Another direction lies in revisiting the way consistency is used for rebalancing scores, for instance by taking more advantage of BMCA computed on multiple values for $c$. Lastly, it is key to understand how this work can be expanded to other types of benchmarks beyond multiple choice, and possibly how these ideas can be used to make LLMs safer in real time, during the inference.

\section{Limitations}
The first limitation of this work is the focus on multiple choice benchmarks only, so the results from this paper do not directly transfer to other types of benchmarks such as open-ended questions. Also, the evaluation comprises a limited set of benchmarks, so further experiments shall be conducted to validate the generalization of our methods. Another limitation is that set of LLMs that we evaluated, given that the size of the models usually tops at around 7B to 8B parameters, so experiments with larger LLMs should also be conducted in the future. Finally, we have not provided any deep discussion on the computational complexity increase of our method, but we decided to not delve into that discussion since, in general terms, the complexity for computing CoRA is roughly equivalent to that of both MCQA+ and MV.

\section{Ethical Statement}
We have not identified any ethical issue, since the LLMs and benchmarks are publicly available and we just followed commonly-used practices.

\bibliography{references}

\clearpage

\appendix

\section{Guessing on Benchmarks}\label{app:guessing_benchmarks}

Requiring consistency of correct answers on multiple, independent answers to the same question of a benchmark is, intuitively, a way to assure that models which are doing, to some extent, random guessing in a multiple choice benchmark. Here we explore the impact of requirement consistency in $M$ trials in terms of determining minimum values for the guessing rate to assure that a given consistency level is met.

We start considering a single multiple-choice question $q$ of $k$ choices which is repeated evaluated by a model $M$ times, yielding answers $llm(q_i), 1 \le i \le M$, where $llm(q_i)=1$ if, and only if, the model produces the correct alternative.

In $M$ trials, the number of possible arrangement of choices where exactly $p$ are correct, $C_M(p)$ is:
\begin{equation}
    C_M(p) = \left( \begin{array}{c} M \\ p \end{array} \right) = \frac{M!}{(M-p)!\,p!}
\end{equation}

Given a guessing rate $r$, where $r=1/p$ if it is a purely random guess, the probability of guessing correctly exactly $p$ of the $M$ trials, $T_r^M(p)$ is:

\begin{equation} 
%\sum_{i=1}^{M}lmm(q_i)=p
    P(T_r^M(p)) = C_M(p)r^p(1-r)^{(M-p)}
\end{equation}

Following, the probability of guessing correctly $p$ or more answers in $M$ trials, $\overline{T}_r^M(p)$ is, clearly:
\begin{equation} 
%\sum_{i=1}^{M}lmm(q_i)=p
    P(\overline{T}_r^M(p)) = \sum_{j=p}^{M}{C_M(j)r^j(1-r)^{(M-j)}}
\end{equation}

The two leftmost columns of table~\ref{app:guessing_benchmarks} show, for different values of $p$, the value of $\overline{T}_r^M(p))$ for $M=10$ trials of $k=5$ choices, when the guessing rate is purely random, $r=1/k$. For instance, the probability of obtaining 10 correct answers in 10 trials of a question if the model is randomly guessing is 0.0000001.

Conversely, now imagine that the model as an ``oracle'' which guesses the correct answer at a certain success rate, $SGR$. We can then compute the minimum success needed to always get at least $p$ correct answers in $M$, what we call the \textit{minimum success guessing rate}, $MSGR(p)$. We computed numerically such values, and the rightmost column of table~\ref{app:guessing_benchmarks} displays the values for $M=10$ trials of $k=5$ choices. It shows, for instance, that that a model has to be guessing at least of a success rate of 0.93 to achieve 6 out of 10 correct answers ($MSGR(6)$), which is equivalent to the requirements of the metric MV proposed in~\citep{pezeshkpour-hruschka-2024-large}.

In our view, a $MSGR(6)= 0.93$ is still insufficient to guarantee that a model actually knows the contents of a multiple-choice benchmark. However, requiring that the model is consistent in 10 out of 10 trials ($MSGR(10)$) warrants that a model can only successfully guess if its success rate is above 0.9999, which we consider a reasonable requirement to consider a model knowledgeable in a subject.

\begin{table}[tb]   
    \centering
    \caption{Guessing probabilities of $p$ or greater correct answers in $M=10$ trials of $k=5$ choices, for random guessing; and minimum success guessing rate $MSGR(p)$.} 
    \label{tab:guessing}
    \begin{tabular}{l|c|c}
    \hline
    \multicolumn{3}{c}{$\overline{T}_r^{10}(p), k=5$} \\
    \hline
    p  & random ($r=1/k$) & $MSGR(p)$ \\
    \hline
    0 & 1.000 & 0.2 \\ 
    1 & 0.893 & 0.54 \\
    2 & 0.624 & 0.66 \\
    3 & 0.322 & 0.75 \\
    4 & 0.121 & 0.82 \\
    5 & 0.033 & 0.88 \\
    6 & 0.006 & 0.93 \\
    7 & 0.0009 & 0.96 \\
    8 & 0.00008 & 0.98 \\
    9 & 0.000004 & 0.99 \\
    10 & 0.0000001 & 0.9999 \\
    \hline

    \end{tabular}
\end{table}

\end{document}